\documentclass[runningheads]{llncs}
\usepackage{graphicx}
\usepackage{amsmath,amssymb} 
\usepackage{color}
\usepackage[width=122mm,left=12mm,paperwidth=146mm,height=193mm,top=12mm,paperheight=217mm]{geometry}
\begin{document}
\pagestyle{headings}
\mainmatter

\title{End-to-End Localization and Ranking for Relative Attributes} 

\titlerunning{End-to-End Localization and Ranking for Relative Attributes}

\authorrunning{Krishna Kumar Singh and Yong Jae Lee}

\author{Krishna Kumar Singh and Yong Jae Lee}


\institute{University of California, Davis}

\maketitle

\begin{abstract}
We propose an end-to-end deep convolutional network to simultaneously localize and rank relative visual attributes, given only weakly-supervised pairwise image comparisons.  Unlike previous methods, our network jointly learns the attribute's features, localization, and ranker.  The localization module of our network discovers the most informative image region for the attribute, which is then used by the ranking module to learn a ranking model of the attribute.  Our end-to-end framework also significantly speeds up processing and is much faster than previous methods.  We show state-of-the-art ranking results on various relative attribute datasets, and our qualitative localization results clearly demonstrate our network's ability to learn meaningful image patches.
\keywords{Relative attributes, Ranking, Localization, Discovery.}
\end{abstract}

\vspace*{-0.05in}
\section{Introduction}
\vspace*{-0.05in}

Visual attributes are mid-level representations that describe semantic properties (e.g., `furry', `natural', `short') of objects and scenes, and have been explored extensively for various applications including zero-shot learning~\cite{palatucci-nips2009,lampert-cvpr2009,parikh-iccv2011}, image retrieval~\cite{siddiqui_CVPR11,kovashka-cvpr2012,kovashka-iccv2013}, fine-grained recognition~\cite{kumar_ICCV09,duan-cvpr2012,zhang-cvpr2014}, and human computer interaction~\cite{branson-eccv2010}.  Attributes have been studied in the binary~\cite{kumar_ICCV09,bourdev-iccv2011}---describing their presence or absence---and relative~\cite{parikh-iccv2011,shrivastava-eccv2012}---describing their relative strength---settings.

Recent work on visual attributes have shown that local representations often lead to better performance compared to global representations~\cite{bourdev-iccv2011,zhang-cvpr2014,sandeep-cvpr2014}.  These methods use pre-trained part detectors to bring the candidate object parts into correspondence to model the attribute, with the assumption that there is at least one well-defined part that corresponds to the attribute.  However, this assumption does not always hold; for example, the exact spatial extent of the attribute {\fontfamily{qcr}\selectfont bald head} can be ambiguous, which means that training a bald head detector itself can be difficult.   Furthermore, since the part detectors are trained independently of the attribute, their learned parts may not necessarily be useful for modeling the attribute. Finally, these methods are designed for binary attributes and are not applicable for relative attributes; however, relative attributes have been shown to be equally or more useful in many settings~\cite{parikh-iccv2011,shrivastava-eccv2012}.

Recently, Xiao and Lee~\cite{xiao-iccv2015} proposed an algorithm that overcomes the above drawbacks by automatically \emph{discovering} the relevant spatial extent of relative attributes given only weakly-supervised (i.e., image-level) pairwise comparisons.  The key idea is to transitively connect ``visual chains'' that localize the same visual concept (e.g., object part) across the attribute spectrum, and then to select the chains that together best model the attribute.  The approach produces state-of-the-art performance for relative attribute ranking.  Despite these qualities, it has three main limitations due to its pipeline nature: 1) The various components of the approach, including the feature learning and ranker, are not optimized jointly, which can lead to sub-optimal performance; 2) It is slow due to time-consuming intermediate modules of the pipeline; 3) In order to build the visual chains, the approach assumes the existence of a visual concept that undergoes a gradual visual change along with the change in attribute strength; however, this does not always hold.  For example, for the  {\fontfamily{qcr}\selectfont natural} attribute for outdoor scenes, there are various visual concepts (e.g., forests and mountains) that are relevant but not consistently present across the images.

\begin{figure}[t!]
    \centering
    \includegraphics[width=0.9\textwidth]{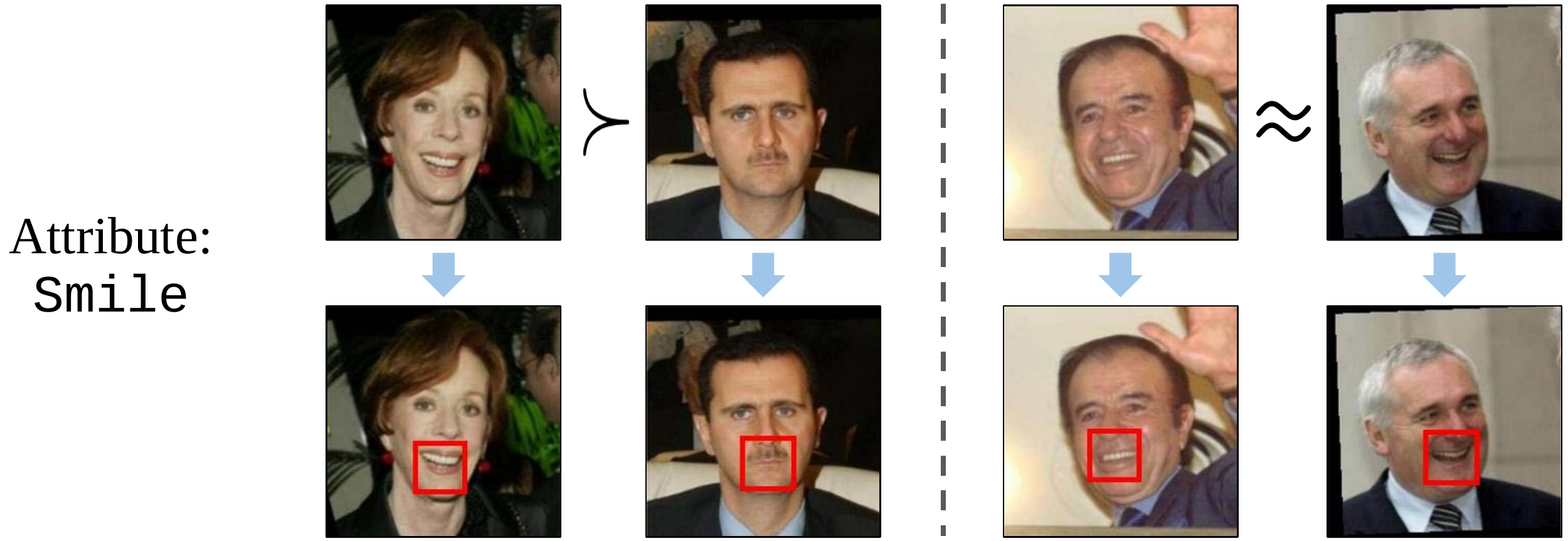}
    \vspace{-0.1in}
    \caption{Given pairwise relative attribute strength comparisons (i.e., greater/less than (left) or similar (right)), our goal is to automatically localize the most informative image regions corresponding to the visual attribute.  For example, the mouth region is the most informative for the attribute {\fontfamily{qcr}\selectfont smile}.  To this end, we train an end-to-end network that discovers the image regions and uses them for relative attribute ranking.}
    \label{fig:teaser}
    \vspace{-0.1in}
\end{figure}

To address these issues, in this paper, we propose an end-to-end deep convolutional network that simultaneously learns to rank and localize relative attributes.  Our setting is the same as in~\cite{xiao-iccv2015}: we are given only weakly-supervised image-level pairwise attribute comparisons, but no supervision on where in each image the attribute is present nor what the attribute looks like.  Thus, the main challenge is to automatically localize the relevant regions in each image pair simultaneously, such that an accurate relative ranking of the attribute for each image pair can be produced.  We tackle this challenge by designing a new architecture that combines a localization network with a ranking network, and optimize a pairwise image ranking loss.  In this way, our approach learns to focus on the regions in each image that are optimal for attribute ranking.  Furthermore, our network is optimized end-to-end, and jointly learns an attribute's features, localization, and ranker, which mutually benefit each other.  This end-to-end framework also significantly speeds up processing.  Finally, unlike~\cite{xiao-iccv2015}, we do not assume that the attribute must be conditioned on the same visual concept across the attribute spectrum.  Instead, our network is free to identify discriminative patterns in each image that are most relevant for attribute ranking (e.g., for the {\fontfamily{qcr}\selectfont natural} attribute, localizing trees for forest images and buildings for city images).  We demonstrate that all of these lead to improved performance over standard pipeline approaches.

Briefly, our method works as follows: we train a Siamese network~\cite{chopra_CVPR05} with a pairwise ranking loss, which takes as input a pair of images and a weak-label that compares the relative strength of an attribute for the image pair.  A Siamese network consists of two identical parallel branches with shared parameters, and during testing, either branch can be used to assign a ranking score to a single image.  Each branch consists of a localization module and a ranking module.  The localization module is modeled with spatial transformer~\cite{Jaderberg-NIPS2015} layers, which discover the most relevant part of the image corresponding to the attribute (see Fig.~\ref{fig:teaser} for examples of localized patches).  The output patch of the localization module is then fed into the ranking module for fine-grained attention for ranking.

\textbf{Contributions.} To our knowledge, this is the first attempt to learn an end-to-end network to rank and localize relative attributes.  To accomplish this, we make two main contributions: (1) a new deep convolutional network that learns to rank attributes given pairwise relative comparisons, and (2) integrating the spatial transformer into our network to discover the image patches that are most relevant to an attribute.  We demonstrate state-of-the-art results on the LFW-10~\cite{sandeep-cvpr2014} face, UT-Zap50K~\cite{yu-cvpr2014} shoe, and OSR~\cite{torralba-ijcv2001} outdoor scene datasets.

\vspace*{-0.05in}
\section{Related Work}
\vspace*{-0.05in}

\textbf{Visual attributes.} Visual attributes serve as an informative and compact representation for visual data.  Earlier work relied on hand-crafted features like SIFT and HOG to model the attributes~\cite{lampert-cvpr2009,farhadi-cvpr2009,kumar-iccv2009,palatucci-nips2009,farhadi-cvpr2010,parikh-iccv2011,rastegari-eccv2012,saleh-cvpr2013,kovashka-iccv2013}.   More recent work use deep convolutional networks to learn the attribute representations, and achieve superior performance~\cite{chung-nipsw2012,escorcia-cvpr2015,zhang-cvpr2014,shankar-arxiv2015}.  While these approaches learn deep representations for \emph{binary} attributes, we instead learn deep representations for \emph{relative} attributes. Concurrent work~\cite{souri_arxiv2015} also trains a deep CNN for the relative setting; however, it does not perform localization as we do.

\textbf{Attribute localization.} Learning attribute models conditioned on local object parts or keypoints have shown to produce superior performance for various recognition tasks~\cite{kumar-eccv2008,bourdev-iccv2011,duan-cvpr2012,sandeep-cvpr2014,zhang-cvpr2014,xiao-iccv2015}.  Most existing work rely on pre-trained part/keypoint detectors or crowd-sourcing to localize the attributes~\cite{kumar-eccv2008,kumar-iccv2009,bourdev-iccv2011,kiapour-eccv2014,zhang-cvpr2014,duan-cvpr2012}. Recently, Xiao and Lee~\cite{xiao-iccv2015} proposed a method to automatically \emph{discover} the spatial extent of relative attributes.  Since it does not rely on pre-trained detectors, it can model attributes that are not clearly tied to object-parts (e.g., {\fontfamily{qcr}\selectfont open} for shoes).  However, the approach consists of several sequential independently-optimized modules in a pipeline system.  As a result, it can be suboptimal and slow.  In contrast, we propose to localize the attributes and train the attribute models \emph{simultaneously} in an end-to-end learning framework.  Similar to~\cite{xiao-iccv2015}, our approach automatically discovers the relevant attribute regions, but is more accurate in ranking and faster since everything is learned jointly.

\textbf{Attention modeling.} Attention models selectively attend to informative locations in the visual data to process at higher resolution.  The key idea is to neglect irrelevant regions (like background clutter) and instead focus on the important regions that are relevant to the task at hand.  Earlier work on bottom-up saliency (e.g.,~\cite{itti-cvpr05,gao-nips2007}) focus on interesting regions in the image, while high-level object proposal methods (e.g.,~\cite{objectness,carreira-mincut,selectivesearch,edgeboxes}) generate candidate object-like regions.  Recent work use deep networks for attention modeling in various tasks including fine-grained classification~\cite{Mnih_NIPS2014,Xiao_CVPR2015}, image caption generation~\cite{xu_arxiv2015}, and image generation~\cite{Gregor_arxiv2015,Tang_NIPS2014}.  In particular, Spatial Transformer Networks~\cite{Jaderberg-NIPS2015} spatially transform the input image to focus on the task-relevant regions, and have shown to improve performance on digit and fine-grained bird recognition.  In our work, we show how to integrate spatial transformer layers into a deep \emph{pairwise ranking} network in order to automatically localize and rank relative attributes for faces, shoes, and outdoor scenes, in the more challenging ranking setting. 

\vspace*{-0.05in}
\section{Approach}
\vspace*{-0.05in}

Given pairs of training images, with each pair ordered according to relative strength of an attribute, our goal is to train a deep convolutional network that learns a function $f:\mathbb{R}^M \rightarrow \mathbb{R}$ to simultaneously discover where in each image the attribute is present and rank each image (with $M$ pixels) according to predicted attribute strength.   Importantly, the only supervision we have are the pairwise image comparisons; i.e., there is no supervision on where in each image the attribute is present nor prior information about the visual appearance of the attribute.  

\vspace*{-0.05in}
\subsection{Input}
\vspace*{-0.05in}

For training, the input to our network is an image pair $(I_1,I_2)$ and a corresponding label $L$ for a given attribute (e.g., {\fontfamily{qcr}\selectfont smile}) indicating whether the image pair belongs to set $E$ or $Q$.  $(I_1,I_2){\in}E$ means that the ground-truth attribute strengths of $I_1$ and $I_2$ are similar to each other, while $(I_1,I_2){\in}Q$ means that the ground-truth attribute strength of $I_1$ is greater than that of $I_2$.  (If attribute strength of $I_1$ is less than that of $I_2$, we simply reorder the two as $(I_2,I_1){\in}Q$.)

For testing, the input is a single image $I_{test}$, and our network uses its learned function $f$ (i.e., network weights) to predict the attribute strength $v=f(I_{test})$.

\begin{figure}[t!]
    \centering
    \includegraphics[width=1\textwidth]{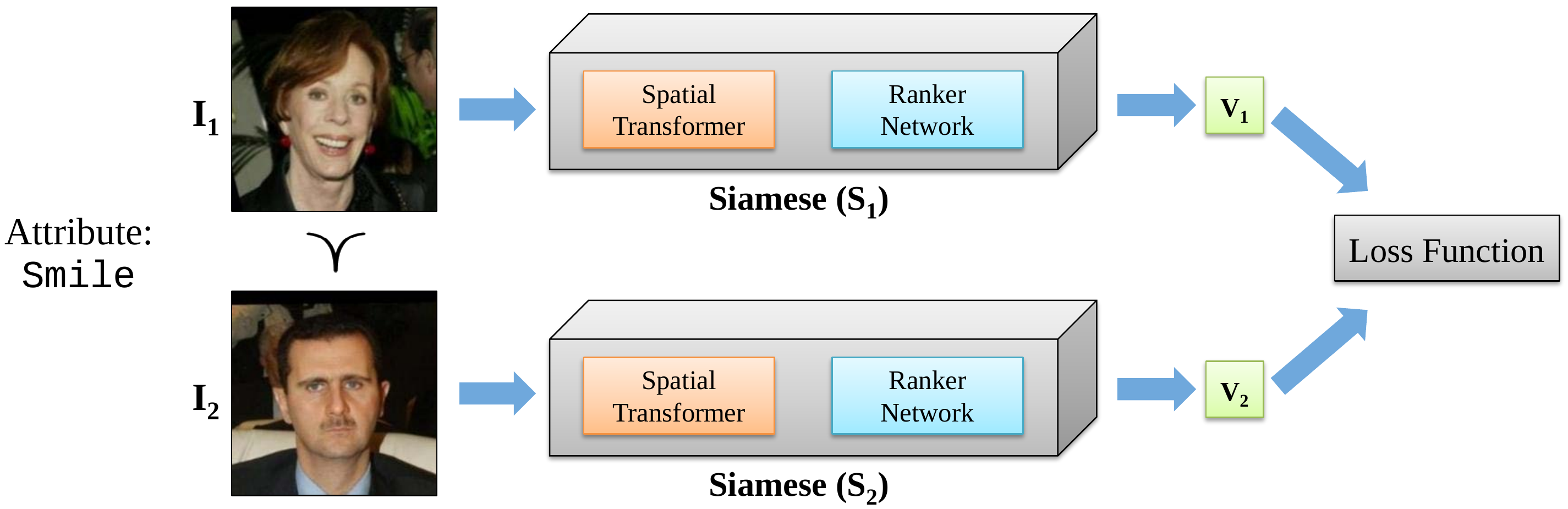}
    \vspace{-0.2in}
    \caption{Overall architecture of our network.  It takes as input a pair of images $(I_1,I_2)$ and a label denoting the images' relative ordering according to ground-truth attribute strength (here, $I_1$ is more {\fontfamily{qcr}\selectfont smiling} than $I_2$).  Each image is fed into a Siamese network, which consists of a Spatial Transformer Network (STN) and a Ranker Network (RN).  The Siamese networks output the predicted attribute scores $v_1$ and $v_2$ for the images, which are used by the loss function to update the network's parameters.}
    \label{fig:cnet}
    \vspace{-0.2in}
\end{figure}

\vspace*{-0.05in}
\subsection{Architecture}
\vspace*{-0.05in}

Fig.~\ref{fig:cnet} shows the overall architecture of our network.  We have a Siamese network~\cite{chopra-CVPR05}, which takes as input an image pair $(I_1, I_2)$ along with its label $L$, and outputs two scalar values $v_1=f(I_1)$ and $v_2=f(I_2)$, which are fed into our loss function.  A Siamese network consists of two identical branches $S_1$ and $S_2$ with shared parameters.  Each branch consists of a {\fontfamily{qcr}\selectfont Spatial Transformer Network} (STN) and a {\fontfamily{qcr}\selectfont Ranker Network} (RN).   The STN is responsible for localizing the relevant image patch corresponding to the visual attribute, while the RN is responsible for generating a scalar value $v$ that denotes the input image's attribute strength.  During testing, only one of the branches is used to produce the attribute strength $v$ for the input test image.

\vspace*{-0.1in}
\subsubsection{Spatial Transformer Network (STN)}

Intuitively, in order to discover the regions in each training image pair that are relevant to the attribute, we could apply a ranking function to various pairs of regions (one region from each image in the pair), and then select the pair that leads to the best agreement with the ground-truth pairwise rankings.  STNs~\cite{Jaderberg-NIPS2015} provide an elegant framework for doing so.  An STN learns an explicit spatial image (or feature map) transformation for each image that is optimal for the task at hand~\cite{Jaderberg-NIPS2015}.   It has two main advantages: (1) fully-differentiable and can be trained with backpropagation; and (2) can learn to translate, crop, rotate, scale, or warp an image without any explicit supervision for the transformation.  By attending to the most relevant image regions, the STN allows the ensuing computation to be dedicated to those regions.  In~\cite{Jaderberg-NIPS2015}, STNs were shown to learn meaningful transformations for digit classification and fine-grained bird classification.

\begin{figure*}[t!]
\centering
    \includegraphics[width=1\textwidth]{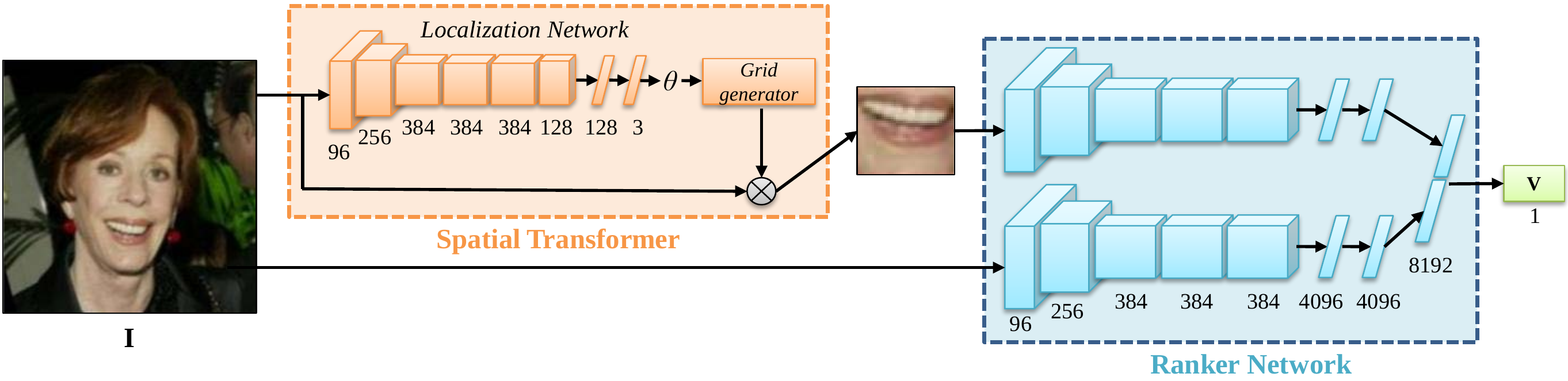}
    \vspace{-0.3in}
    \caption{Illustration of one Siamese network branch.  The input image $I$ goes through the spatial transformer network (STN), which generates the transformation parameters $\theta$.  The transformation is applied to image $I$ to localize the most relevant region corresponding to the attribute.  In this example, the attribute is {\fontfamily{qcr}\selectfont smile}, so the STN localizes the mouth.  Next, the ranker network computes and combines the features of the STN output and image $I$ to compute the attribute strength $v$.}
    \label{fig:STN}
    \vspace*{-0.1in}
\end{figure*}

In this work, we incorporate an STN as part of our end-to-end ranking system, in order to discover the region-of-interest for each relative attribute. The STN's output can then be fed into the ensuing Ranker network, easing its task.  For example, for attribute {\fontfamily{qcr}\selectfont visible-teeth}, it will be easier to optimize the ranking function if the Ranker network receives the mouth region.  While there is no explicit human supervision denoting that the mouth is in fact the most relevant for {\fontfamily{qcr}\selectfont visible-teeth}, the STN can learn to attend to it via its contribution to the ranking loss function.  The architecture of an STN has three main blocks (see Fig.~\ref{fig:STN}, orange box): it first takes the input image and passes it through a convolutional network to obtain the transformation parameters $\theta$, and then a \emph{grid generator} creates a sampling grid, which provides the set of input points that should be sampled to produce the transformed output. With the generated grid and the transformation parameters $\theta$, a bilinear interpolation kernel is applied to sample the input values to produce the output image.  See~\cite{Jaderberg-NIPS2015} for more details.

In our STN, we have three transformation parameters $\theta = [s, t_x, t_y]$, representing isotropic scaling $s$ and horizontal and vertical translation $t_x$ and $t_y$.\footnote{More complex transformations (e.g., affine, thin plate spline) are possible, but we find this transformation to be sufficient for our datasets.}  The transformation is applied as an inverse warp to generate the output image:
\begin{equation}\label{stn}
\vspace*{-0.02in}
\begin{pmatrix}x_i^{in}\\y_i^{in}\end{pmatrix} = \begin{bmatrix} s & ~~0 & ~~t_x \\ 0 & ~~s & ~~t_y  \end{bmatrix} \begin{pmatrix}x_i^{out}\\y_i^{out}\\1\end{pmatrix},
\vspace*{-0.02in}
\end{equation}
where $x_i^{in}$ and $y_i^{in}$ are the input image coordinates, $x_i^{out}$ and $y_i^{out}$ are the output image coordinates, and $i$ indexes the pixels in the output image.

The convolutional network of our STN has six convolutional layers and two fully-connected layers (see Fig.~\ref{fig:STN}, orange box).  The first five layers are equivalent to those of AlexNet~\cite{krizhevsky-nips2012}.  After the max-pooling layer of the 5th convolutional layer (i.e., {\fontfamily{qcr}\selectfont pool5}), we add a convolutional layer consisting of 128 filters of size $ 1 \times 1$ to reduce feature dimensionality.  After the convolutional layer, we add two fully-connected layers; the first takes in 4608 values as input and outputs 128 values, while the second takes the 128 values as input and outputs the final $3$ transformation parameters (i.e., scale, vertical and horizontal translation).  We find these hyperparmeters to provide a good balance of high enough capacity to learn the transformation while minimizing overfitting.

\begin{figure*}[t!]
\centering
    \includegraphics[width=1\textwidth]{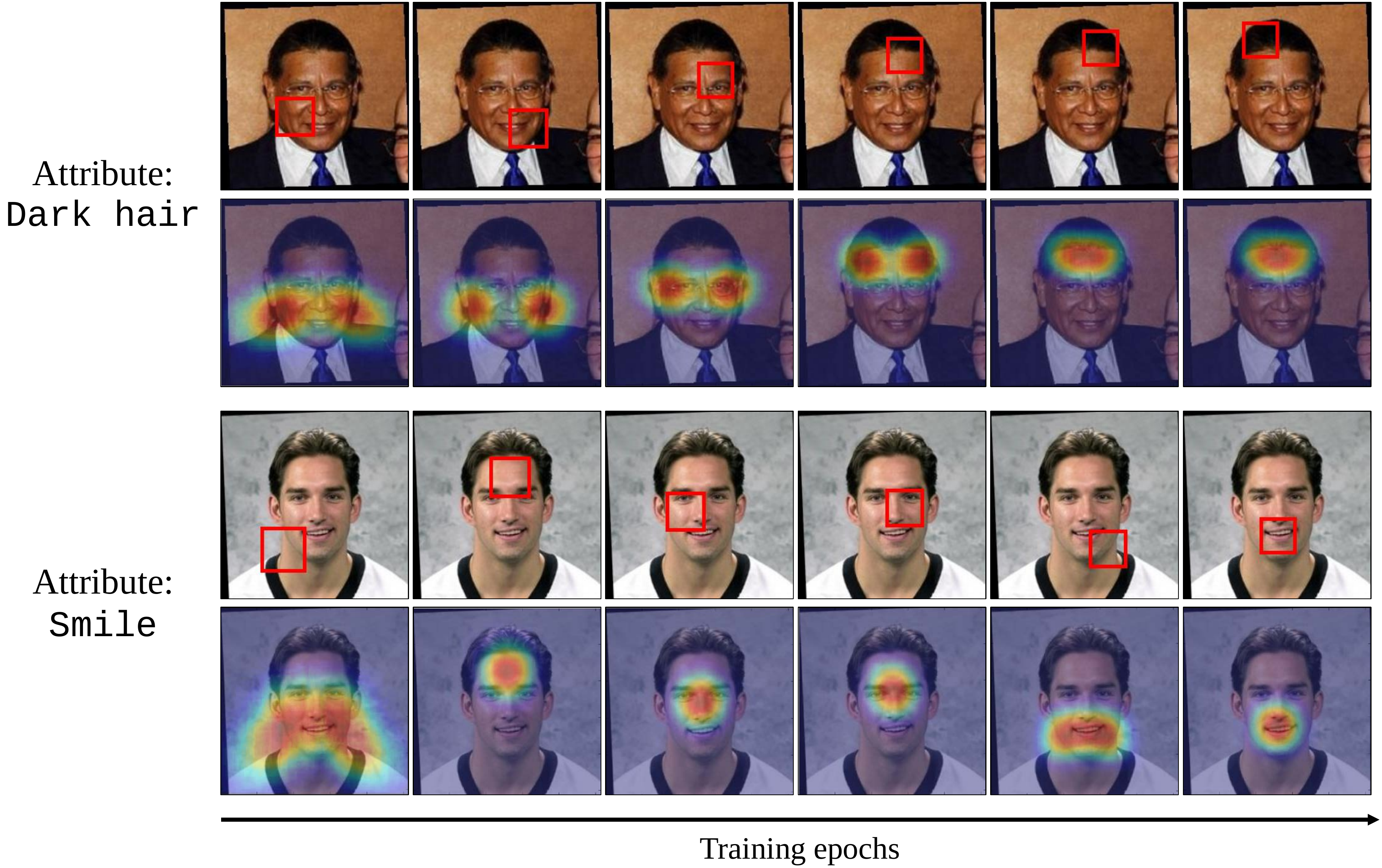}
    \vspace{-0.3in}
    \caption{The localization behavior of the STN during training of our network.  For each attribute, the first row shows the output of the STN for the same image across different training epochs.  The second row shows the distribution of the STN outputs for all the training images across different training epochs (we overlay the heatmap onto an example image).  Notice how the STN output initially has high variance but then gradually converges to the top of the head for {\fontfamily{qcr}\selectfont dark-hair} and the mouth for {\fontfamily{qcr}\selectfont smile}.}
    \label{fig:heatmap}
    \vspace*{-0.1in}
\end{figure*}

Fig.~\ref{fig:heatmap} shows the change in the STN output over the training epochs of our network.  For each attribute, the first row shows an example image with the STN's localized patch in the red box, while the second row shows the distribution of the STN's output over the entire training data overlaid onto the example image (this visualization works because the face images in LFW-10 are roughly aligned).  The STN is initially unsure of the attribute's location and thus has high spatial variance. It then proceeds to search over the various regions in each image and converges to the top of the head for {\fontfamily{qcr}\selectfont dark-hair} and to the mouth for {\fontfamily{qcr}\selectfont smile}.

\vspace*{-0.05in}
\subsubsection{Ranker Network (RN)}

The RN takes the output of the STN (i.e., an image patch) and the original image as input, and computes and combines their features to generate a scalar attribute strength $v$ as output (see Fig.~\ref{fig:STN}, blue box). The key idea of combining the two inputs is to provide the ranker with both the high-resolution image patch that is focused on the visual attribute as well as the entire image to provide global context.  We demonstrate in our experiments that the two sources of information are indeed complementary.

For image pair $(I_1,I_2){\in}E$, the RN will learn to generate similar values for $I_1$ and $I_2$, while for image pair $(I_1,I_2){\in}Q$ the RN will learn to generate a higher value for $I_1$ than $I_2$.  Ultimately, the RN will learn a global ranking function that tries to satisfy all such pairwise constraints in the training data.  Our RN is a Siamese network~\cite{chopra-CVPR05}, with each branch consisting of AlexNet~\cite{krizhevsky-nips2012} without the last fully connected classification layer (i.e., all layers up through {\fontfamily{qcr}\selectfont fc7}).  This generates 4096-D feature vectors for both the image patch and global image, which are concatenated to produce the final 8192-D feature.  A linear layer takes the 8192-D feature to generate a final single value $v$ for the image.  Note that each branch of the RN has shared weights, which reduces the number of parameters by half and helps in reducing overfitting.

\vspace*{-0.05in}
\subsection{Localization and Ranking Loss Function}
\label{sec:lossfcn}
\vspace*{-0.05in}

To learn the parameters of our network, we train it using the loss function introduced in the seminal work of RankNet~\cite{burges-icml05}.  Specifically, we map the outputs $v_1$ and $v_2$ (corresponding to $I_1$ and $I_2$), to a probability $P$ via a logistic function $P= e^{(v_1-v_2)}/(1 + e^{(v_1-v_2)})$, and then optimize the standard cross-entropy loss:
\begin{equation}\label{rankloss}
Rank_{loss}(I_1,I_2) = -L \cdot log(P) - (1-L) \cdot log(1-P),
\end{equation}
where if $(I_1,I_2){\in}Q$ then $L=1$, else if $(I_1,I_2){\in}E$ then $L=0.5$.  This loss function enforces $v_1 > v_2$ when $I_1$ has a higher ground-truth attribute strength than $I_2$, and enforces $v_1 = v_2$ when $I_1$ and $I_2$ have similar ground-truth attribute strengths.  As described in~\cite{burges-icml05}, a nice property of this loss function is that it asymptotes to a linear function, which makes it more robust to noise compared to a quadratic function, and handles input pairs with similar ground-truth attribute strengths in a principled manner as it becomes symmetric with minimum value at 0 when $L=0.5$.

In our initial experiments, we found that large magnitudes for the translation parameters of the STN can lead to its output patch going beyond the input image's boundaries (resulting in a black patch with all 0-valued pixels).  One reason for this is because for any similar pair $(I_1,I_2){\in}E$, its ranking loss can be minimized when the STN produces identical patches for both $I_1$ and $I_2$.  This makes learning difficult because the resulting gradient direction of the ranking loss with respect to the transformation parameters becomes uninformative, since the same black patch will be produced in all nearby spatial directions.  To handle this, we introduce a simple loss that updates the transformation parameters to bring the STN's output patch back within the image boundaries if it goes outside:
\begin{equation}\label{stloss}
ST_{loss}(I) = (C_x - s \cdot t_x)^2 + (C_y - s \cdot t_y)^2,
\end{equation}
where $t_x$ and $t_y$ are horizontal and vertical translation, respectively, $s$ is isotropic scaling, and $C_x$ and $C_y$ are the center $x$ and $y$ pixel-coordinates of the input image $I$, respectively. $ST_{loss}$ is simply the squared distance of the center coordinates of the output patch from the input image's center, and is differentiable.  The loss increases as $(s \cdot t_x,s \cdot t_y)$ moves farther away from the image center, so the output patch will be forced to move back toward the image.  Importantly, we do not apply this loss if the output patch's center coordinates are within the image's boundaries (i.e., this loss does not bias the STN to produce regions that are near the image center).

Putting Eqns.~\ref{rankloss} and~\ref{stloss} together, our final loss function is:
\begin{equation}\label{loss}
\hspace*{-0.07in}
Loss = \frac{1}{N} \sum_i (1-\lambda_1^i)(1-\lambda_2^i) \cdot Rank_{loss}(I_1^i,I_2^i) + \lambda_1^i \cdot ST_{loss}(I_1^i) + \lambda_2^i \cdot ST_{loss}(I_2^i),
\end{equation}
where $N$ is the total number of training image pairs, $i$ indexes over the training pairs, $\lambda_1^i$=$1$ ($\lambda_2^i$=$1$) if the center coordinates of the STN's output patch of $I_1^i$ ($I_2^i$) falls outside of the image's boundaries.  We optimize Eqn.~\ref{loss} with backpropagation to learn the entire network's weights $f$.  Note that the gradient computed for $ST_{loss}$ is only backpropagated through the STN and does not affect the RN.  If both $\lambda_1^i$=$\lambda_2^i$=$0$ (the STN's output patches for both $I_1^i$ and $I_2^i$ are within their image boundaries), then the gradient computed for $Rank_{loss}$ is backpropagated through the entire network (i.e., both the RN and STN are updated).

\vspace*{-0.05in}
\section{Results}
\vspace*{-0.05in}

In this section, we analyze our network's attribute localization and ranking accuracy through both qualitative and quantitative results.

\vspace*{-0.1in}
\subsubsection{Datasets.}  LFW-10~\cite{sandeep-cvpr2014}: It consists of 10 face attributes ({\fontfamily{qcr}\selectfont bald-head}, {\fontfamily{qcr}\selectfont dark-hair}, {\fontfamily{qcr}\selectfont eyes-open}, {\fontfamily{qcr}\selectfont good-looking}, {\fontfamily{qcr}\selectfont masculine-looking}, {\fontfamily{qcr}\selectfont mouth-open}, {\fontfamily{qcr}\selectfont smile}, {\fontfamily{qcr}\selectfont visible-teeth}, {\fontfamily{qcr}\selectfont visible-forehead}, and {\fontfamily{qcr}\selectfont young}). There are 1000 training images and 1000 test images, with 500 pairs per attribute for both training and testing.  We use the same train-test split used in~\cite{sandeep-cvpr2014}.

UT-Zap50K-1~\cite{yu-cvpr2014}: It consists of 4 shoe attributes ({\fontfamily{qcr}\selectfont open}, {\fontfamily{qcr}\selectfont pointy}, {\fontfamily{qcr}\selectfont sporty} and {\fontfamily{qcr}\selectfont comfort}). There are 50,025 shoe images, and 1388 training and 300 testing pairs per attribute.  We use the train-test splits provided by~\cite{yu-cvpr2014}.

OSR~\cite{torralba-ijcv2001}: It consists of  outdoor scene attributes ({\fontfamily{qcr}\selectfont natural}, {\fontfamily{qcr}\selectfont open}, {\fontfamily{qcr}\selectfont perspective}, {\fontfamily{qcr}\selectfont large-objects}, {\fontfamily{qcr}\selectfont diagonal-plane}, and {\fontfamily{qcr}\selectfont close-depth}).  There are 2688 images, and we use the same train/test split as in~\cite{parikh-iccv2011,yu-cvpr2014}.

\vspace*{-0.1in}
\subsubsection{Implementation details.}

We train a separate network for each visual attribute.  We initialize the STN and RN weights with those of AlexNet~\cite{krizhevsky-nips2012} pre-trained on ImageNet classification up through {\fontfamily{qcr}\selectfont conv5} and {\fontfamily{qcr}\selectfont fc7}, respectively.  We first train our network without the global image; i.e., the RN only receives the output of the STN to generate the attribute score $v$.  We then retrain the entire network---in which we use both the STN output and the global image to train the RN---with the STN weights initialized with the initially learned weights.  We find that this setup helps the STN localize the attribute more accurately, since initially the STN is forced to find the optimal image region without being able to rely on the global image when predicting the attribute strength.

\begin{figure}[p!]
    \centering
    \includegraphics[width=1\textwidth]{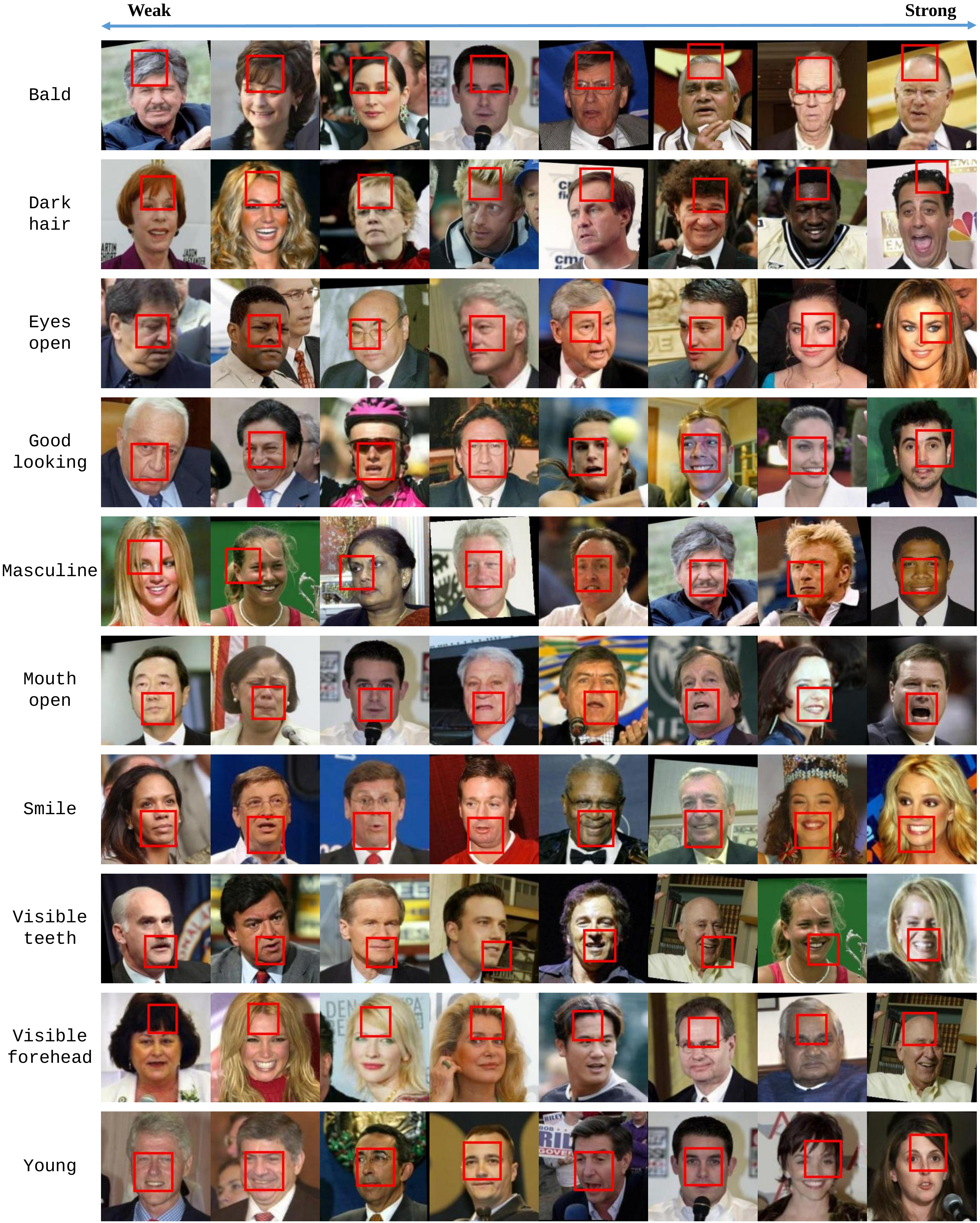}
    \vspace{-0.15in}
    \caption{Qualitative results on LFW-10 test images.  Each row corresponds to an attribute, with the images uniformly sampled according to predicted attribute strength. In each image, the STN localization is depicted in the red box.  It corresponds to meaningful regions for each localizable attribute (e.g., top of the head for {\fontfamily{qcr}\selectfont bald-head} and {\fontfamily{qcr}\selectfont dark-hair}; forehead for {\fontfamily{qcr}\selectfont visible-forehead}; mouth for {\fontfamily{qcr}\selectfont mouth-open}, {\fontfamily{qcr}\selectfont smile} and {\fontfamily{qcr}\selectfont visible-teeth}; eyes for {\fontfamily{qcr}\selectfont eyes-open}). For more global attributes like {\fontfamily{qcr}\selectfont good-looking}, {\fontfamily{qcr}\selectfont masculine-looking}, and {\fontfamily{qcr}\selectfont young}, there is no definite answer, but our method tends to focus on larger areas that encompass the eyes, nose, and mouth.  Finally, the ranking obtained by our method is accurate for all attributes.}
    \label{fig:face_qualitative}
    \vspace{-0.1in}
\end{figure}

For training, we use a mini-batch size of 25 image pairs for SGD, and train the network for 400, 200, and 15 epochs for LFW-10, UT-Zap50K, and OSR, respectively. We set the learning rate for the RN and STN to be 0.001 and 0.0001, respectively, and fix momentum to 0.9.  Also, we set the relative learning rate for the scale parameter to be one-tenth of that of the translation parameters, as we find scaling to be more sensitive and can transform the image drastically. We initialize the scale to be 1/3 of the image size for LFW-10 and 1/2 for UT-Zap50K and OSR, based on initial qualitative observations. Translation is initialized randomly for all datasets.  We use random crops of size $227 \times 227$ from our $256 \times 256$ input image during training, and average the scores for 10 crops (4 corners plus center, and same with horizontal flip) during testing.

\vspace*{-0.1in}
\subsubsection{Baselines.}

We compare against the state-of-the art method of Xiao and Lee~\cite{xiao-iccv2015}, which uses a pipeline system to discover the spatial extent of relative attributes and trains an SVM ranker to rank the images.  We report the method's results obtained by combining the features computed over the global image and discovered patches, which produce the best accuracy.  

\begin{figure}[t!]
    \centering
    \includegraphics[width=1\textwidth]{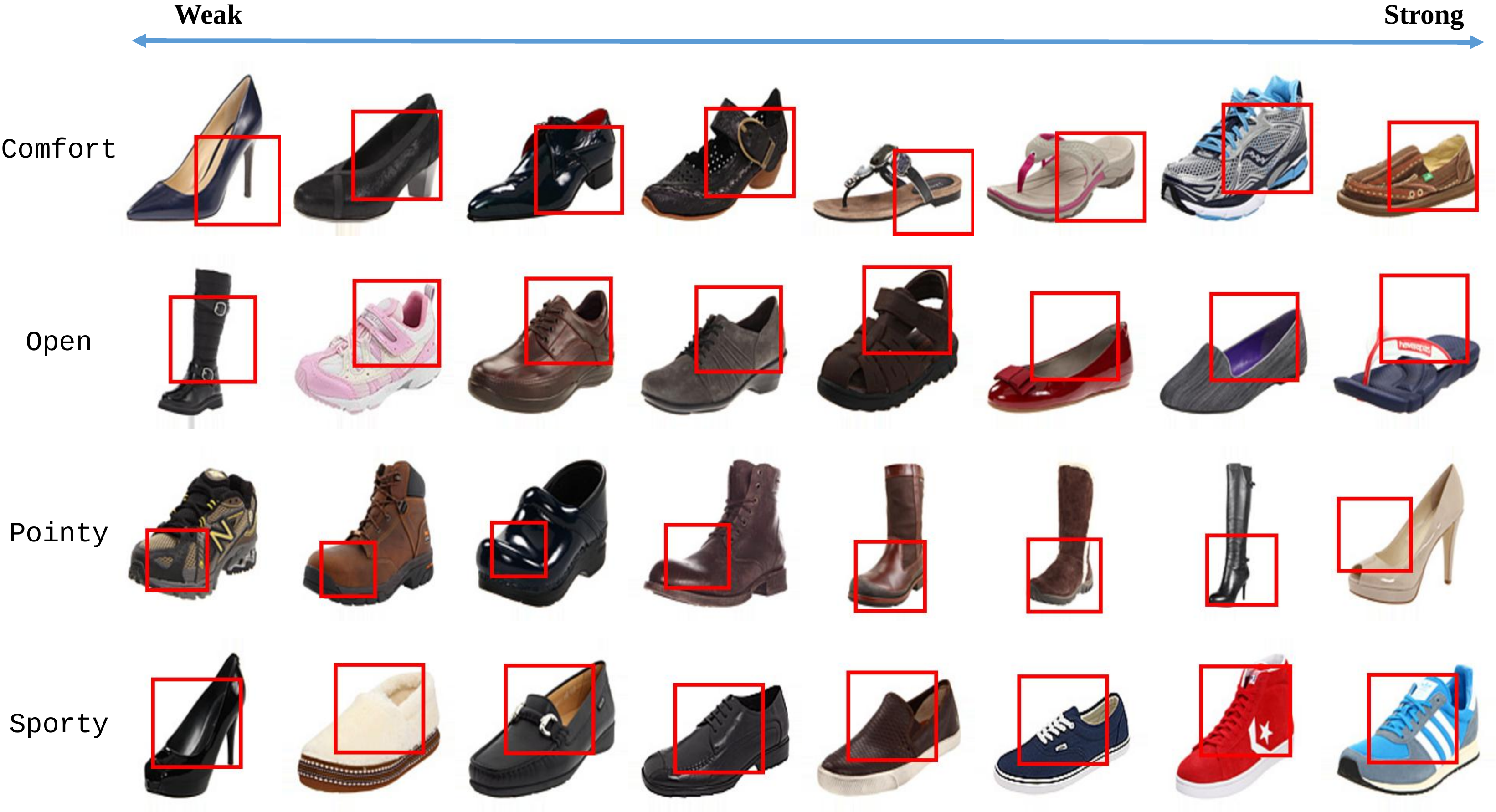}
    \vspace{-0.2in}
    \caption{Qualitative results on UT-Zap50K-1 test images.  The STN localizes the relevant image regions: toe end for {\fontfamily{qcr}\selectfont pointy}, heel for {\fontfamily{qcr}\selectfont comfort}, top opening for {\fontfamily{qcr}\selectfont open}, and area around the laces for {\fontfamily{qcr}\selectfont sporty}.  Our method's ranking is also accurate for each attribute.}
    \label{fig:shoe_qualitative}
    \vspace{-0.1in}
\end{figure}

We also compare against~\cite{sandeep-cvpr2014}, which computes dense SIFT features on keypoints detected using a supervised facial keypoint detector~\cite{zhu-cvpr2012} to train an SVM ranker, and~\cite{parikh-iccv2011}, which trains an SVM ranker with global image features.  For~\cite{parikh-iccv2011}, we compare against the results reported using CNN (pre-trained AlexNet {\fontfamily{qcr}\selectfont pool5}) features in~\cite{xiao-iccv2015}.  Finally, we compare against the local learning method of~\cite{yu-cvpr2014}.  We report its numbers generated using GIST+color-histogram features.

\vspace*{-0.05in}
\subsection{Qualitative results of localized attributes}

We first visualize our attribute localization results.  In Fig.~\ref{fig:face_qualitative}, we show the results for the face attributes on the LFW-10 test images.  Each row corresponds to a face attribute, and the red box in each image indicates the output of the STN.  The images in each row are uniformly sampled after sorting them according to the attribute strength predicted by our network.  We can see that our network localizes the relevant regions for the various face attributes.  For example, it localizes the mouth region for {\fontfamily{qcr}\selectfont mouth-open}, {\fontfamily{qcr}\selectfont smile}, and {\fontfamily{qcr}\selectfont visible-teeth}; the top of the head for {\fontfamily{qcr}\selectfont bald-head} and {\fontfamily{qcr}\selectfont dark-hair}; near the eyes for {\fontfamily{qcr}\selectfont eyes-open}; and the forehead for {\fontfamily{qcr}\selectfont visible-forehead}.  For more global attributes like {\fontfamily{qcr}\selectfont good-looking}, {\fontfamily{qcr}\selectfont masculine-looking}, and {\fontfamily{qcr}\selectfont young}, there is no definite answer but our network tends to localize larger portions of the face.

In Fig.~\ref{fig:shoe_qualitative}, we show the results for the shoe attributes on the UT-Zap50K-1 test images.  Again, our network localizes the relevant regions for the different shoe attributes.  It localizes the heel for {\fontfamily{qcr}\selectfont comfort}; the toe end for {\fontfamily{qcr}\selectfont pointy}; and the top opening for {\fontfamily{qcr}\selectfont open}.  Finally, our network is able to produce accurate image rankings using the localized image parts, as shown in both Fig.~\ref{fig:face_qualitative} and Fig.~\ref{fig:shoe_qualitative}.  For example, we can see the progression of light-to-dark hair for {\fontfamily{qcr}\selectfont dark-hair}, and closed-to-open shoes for {\fontfamily{qcr}\selectfont open}.  

Overall, these qualitative results demonstrate that our method is able to produce accurate localizations.

\vspace*{-0.05in}
\subsection{Quantitative results for attribute ranking}
\vspace*{-0.05in}

We next evaluate quantitative ranking accuracy.  We report the percentage of test image pairs whose relative attribute ranking is predicted correctly.

Table~\ref{table:LFW10Results} shows the results on LFW-10.  First, the baseline method of~\cite{parikh-iccv2011} uses a global representation instead of a local one to model the attributes and produces the lowest accuracy.  The state-of-the-art method of Xiao and Lee~\cite{xiao-iccv2015} outperforms the baseline method of~\cite{sandeep-cvpr2014} because it automatically discovers the relevant regions of an attribute without relying on pretrained keypoint detectors whose detected parts may be irrelevant to the attribute.  Still, across all attributes, we improve on average by  $2.25\%$ absolute over the method of Xiao and Lee~\cite{xiao-iccv2015}, and produce the best results on seven attributes.  This shows the benefit of our end-to-end network, which learns the attribute features, localization, and ranker jointly.  In contrast, the method of Xiao and Lee~\cite{xiao-iccv2015} optimizes each step independently, which leads to sub-optimal results.

\begin{table*}[t]
	\begin{center}
        \hspace*{-0.12in}
		\scriptsize
		\begin{tabular}{ | c | c c c c c c c c c c | c |}
			\hline
			& BH & DH & EO & GL & ML & MO & S & VT & VF & Y & Mean \\			
			\hline
			Parikh \& Grauman~\cite{parikh-iccv2011}+CNN & 78.10 & 83.09 & 71.43 & 68.73 & 95.40 & 65.77 & 63.84 & 66.46 & 81.25 & 72.07 & 74.61\\
			Sandeep et al.~\cite{sandeep-cvpr2014} & 82.04 & 80.56 & 83.52 & 68.98 & 90.94 & 82.04 & 85.01 & 82.63 & 83.52 & 71.36	& 81.06 \\
			Xiao \& Lee~\cite{xiao-iccv2015} & 83.21 & 88.13	& 82.71 & \textbf{72.76} & 93.68 & 88.26 & \textbf{86.16} & 86.46 & \textbf{90.23} & 75.05 & 84.66\\
			Ours & \textbf{83.94} & \textbf{92.58}	& \textbf{90.23} & 71.21 & \textbf{96.55} & \textbf{91.28} & 84.75 & \textbf{89.85} & 87.89 & \textbf{80.81} & \textbf{86.91}\\
			\hline
		\end{tabular}
		\caption{Attribute ranking accuracy on LFW-10.  On average, we outperform all previous methods, and achieve the best accuracy for 7 out of 10 attributes.  These results show the advantage of our end-to-end network, which simultaneously learns to localize and rank relative attributes.}
		\label{table:LFW10Results}
	\end{center}
	\vspace*{-0.3in}
\end{table*}

\begin{table}[t]
	\begin{center}	
		\scriptsize
		{
			\begin{tabular}{ | c | c  c  c  c | c |}
				\hline
				& Open & Pointy & Sporty & Comfort & Mean \\ \hline				
				Parikh and Grauman~\cite{parikh-iccv2011}+CNN & 94.37 & 93.97 & 95.40 & 95.03 & 94.69 \\
                Yu and Grauman~\cite{yu-cvpr2014} & 90.67 & 90.83 & 92.67 & 92.37 & 91.64 \\
				Xiao and Lee~\cite{xiao-iccv2015} & \textbf{95.03} & 94.80 & 96.47 & 95.60	& 95.47 \\
				Ours & 94.87 & \textbf{94.93} & \textbf{97.47} & \textbf{95.87}	& \textbf{95.78} \\
				\hline
			\end{tabular}
		}
		\caption{Attribute ranking accuracy on UT-Zap50K-1. The shoe images are well-aligned, centered, and have clear backgrounds, so all methods obtain very high accuracy and results are nearly saturated.}
		\label{table:utzap50kResults}
	\end{center}
	\vspace*{-0.3in}
\end{table}

\begin{table}[t]
	\begin{center} {
			\scriptsize 
			\begin{tabular}{| c | c  c  c  | c |}
				\hline
				& Open & Pointy & Sporty & Mean \\
                \hline				
				Parikh \& Grauman~\cite{parikh-iccv2011}+CNN 	& 77.10				& 72.50				& 71.56 			 	& 73.72				\\
				Xiao \& Lee~\cite{xiao-iccv2015} 						& 80.15				& \textbf{82.50}				& 88.07				 	& 83.58				\\
				Ours 	& \textbf{89.31}			& \textbf{82.50}				& \textbf{93.58}			 	& \textbf{88.46}				\\ \hline
			\end{tabular}
		}
		\caption{Attribute ranking accuracy on Shoe-with-Attribute using the models trained on UT-Zap50K-1.  We significantly outperform the previous state-of-the-art (Xiao and Lee~\cite{xiao-iccv2015}) on this more challenging dataset, which demonstrates our network's cross-dataset generalization ability.}
		\label{table:crossResults}
	\end{center}
	\vspace{-0.3in}
\end{table}

Table~\ref{table:utzap50kResults} shows the results on UT-Zap50K-1.  For this dataset, our improvement over the baselines is marginal because the shoe images are so well-aligned, centered, and have clear backgrounds.  Consequently, all baselines obtain very high accuracy and the results are nearly saturated.  Thus, following~\cite{xiao-iccv2015}, we perform a cross-dataset experiment on the more challenging Shoes-with-Attribute~\cite{kovashka-cvpr2012} dataset, whose shoe images are not as well-aligned and have more variation in style and scale.  Shoes-with-Attribute has three overlapping attributes with UT-Zap50K-1 ({\fontfamily{qcr}\selectfont open}, {\fontfamily{qcr}\selectfont pointy}, and {\fontfamily{qcr}\selectfont sporty}) with 140 annotated image pairs per attribute.  We take our models trained on UT-Zap50K-1, and test on Shoes-with-Attribute in order to evaluate cross-dataset generalization ability.  Table~\ref{table:crossResults} shows the results.  We get a significant boost of $4.88\%$ absolute over the method of Xiao and Lee~\cite{xiao-iccv2015}. This demonstrates that our joint training of the attribute features, localization, and ranker lead to more robustness to appearance and scale variations, compared to training them independently.

Finally, Table~\ref{table:scene} shows the results on OSR, which contains very different looking images for the same relative attribute (e.g., forest vs. city images for {\fontfamily{qcr}\selectfont natural}). We compare against the previous state-of-the-art local learning method of Yu and Grauman~\cite{yu-cvpr2014}, the relative forest method of~\cite{li_ACCV12}, and global image-based method of~\cite{parikh-iccv2011}.  We obtain an overall accuracy of 97.02\%, which is better than all baselines.  More importantly, this result shows our advantage over the method of Xiao and Lee~\cite{xiao-iccv2015}, which is not applicable to this dataset, since it requires a visually consistent concept (e.g., the same object) for an attribute to build its visual chains.  In contrast, our method can handle drastically different visual concepts for the same attribute (e.g., forest image with trees being more {\fontfamily{qcr}\selectfont natural} than city image with buildings) since the STN only needs to localize the most relevant region in each image without requiring that the concept be present in other images.

\begin{table}[t]
	\begin{center} {
			\scriptsize 
			\begin{tabular}{| c | c  c  c  c  c  c | c |}
				\hline
				& Natural & Open & Perspective & Size-Large & Diagonal & Depth-Close  & Mean \\
                \hline				
				Parikh \& Grauman~\cite{parikh-iccv2011} 	& 95.03	& 90.77	& 86.73	& 86.23	& 86.50	& 87.53	& 88.80				\\
				Parikh \& Grauman~\cite{parikh-iccv2011} + CNN	& 98.02	& 94.52	& 93.04	& 94.04	& 95.00	& 95.25	& 94.98				\\
				Li et al.~\cite{li_ACCV12} 			        & 95.24	& 92.39	& 87.58	& 88.34	& 89.34	& 89.54	& 90.41			\\
				Yu and Grauman~\cite{yu-cvpr2014}		& 95.70	& 94.10	& 90.43	& 91.10	& 92.43	& 90.47	& 92.37	        			\\
				Ours 	& \textbf{98.89}	& \textbf{97.20}	& \textbf{96.31}	& \textbf{95.98}	& \textbf{97.64}	& \textbf{96.10}	& \textbf{97.02}				\\ \hline
			\end{tabular}
		}
		\caption{Attribute ranking accuracy on OSR.  We outperform previous methods.}
		\label{table:scene}
	\end{center}
	\vspace{-0.3in}
\end{table}

\vspace*{-0.05in}
\subsection{Ablation study}
\vspace*{-0.05in}

We study the contribution that the global image versus the output region of the STN has on ranking performance.  For this, we train and compare two baseline networks: (1) the RN is trained with only the global image as input (\textbf{Global image}), i.e., we do not have the STN as part of the network; and (2) the RN is trained with only the STN output region as input without the global image (\textbf{STN output}).  For both baseline networks, the final linear layer uses the newly-learned 4096-D feature output of the RN to generate the attribute strength $v$.

Table~\ref{table:Ablation_study} shows quantitative ranking accuracy of these baseline networks on LFW-10.  Overall, the STN output baseline outperforms the Global image baseline.  It performs especially well for attributes that are conditioned on small facial parts like {\fontfamily{qcr}\selectfont eyes-open}, {\fontfamily{qcr}\selectfont mouth-open}, and {\fontfamily{qcr}\selectfont visible-teeth}.  This is mainly because the Global image baseline needs to process the entire input image, which means that small object parts like the eyes or mouth have very low resolution, and thus, cannot receive the fine-grained attention that they need.  In contrast, the STN can attend to those small informative parts, and provide a high-resolution (cropped-out) image for the RN to learn the ranking function.  Finally, the third row in Table~\ref{table:Ablation_study} shows the result of our full model, i.e., combining both the global input image and STN output to train the RN.  Our full model produces the best accuracy for eight out of the 10 attributes, which shows that the global contextual information from the input image and the fine-grained information from the localized STN output are complementary.  The boost is especially significant for mid-sized and global attributes like {\fontfamily{qcr}\selectfont dark-hair}, {\fontfamily{qcr}\selectfont visible-forehead}, and {\fontfamily{qcr}\selectfont young}, likely because they require both fine-grained information of specific parts as well as more global contextual information of the entire face.

\begin{table*}[t!]
	\begin{center}
		\scriptsize
		\begin{tabular}{| c | c c c c c c c c c c | c |}
			\hline
			& BH & DH & EO & GL & ML & MO & S & VT & VF & Y & Mean \\
			\hline
			Global image & \textbf{84.31} &	90.21 &	82.71	& 69.97	& 94.83	& 80.20	 & 80.79	& 79.38	& 85.55	& 77.19	& 82.51 \\
			STN output & 78.10	& 89.32	& \textbf{93.23}	& 66.56 &	95.4	& \textbf{91.28} & 	84.46	& 88.62	& 85.55	& 73.77	& 84.63
			\\					
			Combined & 83.94 & \textbf{92.58}	& 90.23 & \textbf{71.21} & \textbf{96.55} & \textbf{91.28} & \textbf{84.75} & \textbf{89.85} & \textbf{87.89} & \textbf{80.81} & \textbf{86.91}\\
			\hline
		\end{tabular}
		\caption{Ablation study on LFW-10 comparing the contribution of the global input image (1st row) and the STN's localized output region (2nd row) for attribute ranking.  Our combined model (3rd row) produces the best performance, showing that the information present in the global image and STN output are complementary.}
		\label{table:Ablation_study}
	\end{center}
	\vspace*{-0.3in}
\end{table*}

\vspace*{-0.05in}
\subsection{Computational speed analysis}\label{runtime}
\vspace*{-0.05in}

Finally, we analyze the computational speed of our network.  Our approach is significantly faster than the related baseline method of Xiao and Lee~\cite{xiao-iccv2015}. Since that method has to process a sequence of time-consuming modules including feature extraction, nearest neighbor matching, iterative SVM classifier training to build the visual chains, and training an SVM ranker to rank the chains, it takes $\sim$10 hours to train one attribute model on LFW-10 using a cluster of 20 CPU nodes with 2 cores each. In contrast, our end-to-end network only takes $\sim$3 hours to train one attribute model using a single Titan X GPU. For testing, our approach takes only 0.011 seconds per image compared to 1.1 seconds per image for \cite{xiao-iccv2015}.
\vspace*{-0.05in}
\section{Discussion}
\vspace*{-0.05in}

We presented a novel end-to-end deep network that combines a localization module with a ranking module to jointly localize and rank relative attributes.  Our qualitative results showed our network's ability to accurately localize the meaningful image patches corresponding to an attribute. We demonstrated state-of-the-art attribute ranking performance on benchmark face, shoe, and outdoor scene datasets. One limitation of our approach is that it can only localize  one image part for a visual attribute. However, for certain attributes there can be multiple relevant parts. We would like to explore this issue further either by having multiple spatial transformers or directly predicting pixel-level relevance.

\vspace{-3pt}
\paragraph{\textbf{Acknowledgements.}} This work was supported in part by an Amazon Web Services Education Research Grant and GPUs donated by NVIDIA.


\end{document}